%% file: main.tex
\documentclass[10pt,twocolumn,letterpaper]{article}
\usepackage[table,dvipsnames]{xcolor} %
\usepackage{iccv}              %
\usepackage{booktabs}
\usepackage{multirow}
\usepackage{graphicx} %
\usepackage{colortbl}

\input{preamble}

\definecolor{iccvblue}{rgb}{0.21,0.49,0.74}
\usepackage[pagebackref,breaklinks,colorlinks,allcolors=iccvblue]{hyperref}

\def\confName{ICCV}
\def\confYear{2025}

\title{Can Video Diffusion Model Reconstruct 4D Geometry ?}

\author{
Jinjie Mai \authorsep Wenxuan Zhu \authorsep Haozhe Liu \authorsep Bing Li  \authorsep Cheng Zheng \authorsep Jürgen Schmidhuber \authorsep Bernard Ghanem \\
\normalsize King Abdullah University of Science and Technology (KAUST) 
\\
{\tt\small \{jinjie.mai,bernard.ghanem\}@kaust.edu.sa} \\
\href{https://wayne-mai.github.io/publication/sora3r_arxiv_2025/}{\texttt{\textcolor{purple}{\underline{Project Page}}}}
}

\begin{document}
\maketitle

\input{sec/0_abstract}
\input{sec/1_intro}

\input{sec/2_related_works}

\input{sec/4_method}

\input{sec/5_experiment}
\input{sec/5_result}

\input{sec/6_ablations}
\input{sec/7_conclusion}

\input{sec/8_acknowledgement}

\clearpage

{
    \small
    \bibliographystyle{ieeenat_fullname}
    \bibliography{main}
}

\end{document}

%% file: preamble.tex
\definecolor{mycolor1}{HTML}{1F77B4} %
\definecolor{mycolor2}{HTML}{FF7F0E} %
\definecolor{mycolor3}{HTML}{2CA02C} %

\newcommand{\authorsep}{\hspace{8pt}}

\newcommand{\mysection}[1]{\vspace{3pt}\noindent\textbf{#1.}}

%% file: sec/0_abstract.tex
\begin{abstract}
Reconstructing dynamic 3D scenes (i.e., 4D geometry) from monocular video is an important yet challenging problem. 
Conventional multiview geometry-based approaches often struggle with dynamic motion, whereas recent learning-based methods either require specialized 4D representation or sophisticated optimization. 
In this paper, we present \emph{Sora3R}, a novel framework that taps into the rich spatiotemporal priors of large-scale video diffusion models to directly infer 4D pointmaps from casual videos. 
\emph{Sora3R} follows a two-stage pipeline:
(1)~we adapt a \emph{pointmap VAE} from a pretrained video VAE, ensuring compatibility between the geometry and video latent spaces; 
(2)~we finetune a diffusion backbone in combined video and pointmap latent space to generate coherent 4D pointmaps for every frame. 
\emph{Sora3R} operates in a fully feedforward manner, requiring no external modules (e.g., depth, optical flow, or segmentation) or iterative global alignment. 
Extensive experiments demonstrate that \emph{Sora3R} reliably recovers both camera poses and detailed scene geometry, achieving performance on par with state-of-the-art methods for dynamic 4D reconstruction across diverse scenarios.
\end{abstract}

%% file: sec/1_intro.tex
\section{Introduction}

\input{figures/fig_train}

\noindent The ability to capture and reconstruct detailed 3D structures from visual data has long been a cornerstone of computer vision, powering critical tasks in robotics, autonomous driving, augmented reality (AR), and virtual reality (VR). Traditional multiview geometry-based frameworks~\cite{modern_ba,mvg_vgg} for Simultaneous Localization and Mapping (SLAM), Structure-from-Motion (SfM), and 3D reconstruction have matured into robust pipelines like COLMAP~\cite{colmap} and ORB-SLAM~\cite{orb_slam,orb_slam3}, offering reliable camera pose estimation together with sparse or semi-dense maps of static scenes. Despite their enduring success, these methods often struggle with dynamic objects or scenes, prompting the community to filter out dynamic and non-rigid components through motion segmentation~\cite{romo_sfm,dynaslam} or other heuristics~\cite{leapvo, casualsam, particlesfm}. As the demand for video-based understanding grows, there is increasing interest in not only  achieving\emph{dense} geometric reconstructions but also advancing toward \emph{4D} \textit{(3D + temporal 1D)} modeling --- capturing both spatial structure and temporal dynamics within a scene.

Such 4D reconstruction is critical for many real-world applications abound: robotics platforms~\cite{orbit_isaac} increasingly train in virtual environments, AR/VR~\cite{aria_glass} users seek to teleport dynamic real-world scenes into digital spaces, and content creators aspire to manipulate 4D assets for visual effects or physical consistent interaction~\cite{Genesis}.
While existing research has progressed considerably, many approaches either (a)~\cite{dnerf, 4dgs, reconx, viewcrafter, splinegs, nutworld, bullet_time, flare} reconstruct various 4D representations like 4D NeRF~\cite{nerf} or 4D Gaussian Splatting~\cite{3dgs} from video synthesis; and (b)~\cite{casualsam,megasam,leapvo,robustcvd} learn from geometry supervision signals like depth, optical flow, point tracks, or camera pose.

Recently, DUSt3R~\cite{dust3r} opened a new chapter by introducing pointmap regression from pairwise images,
demonstrating a promising way forward for dense 3D reconstruction: each pixel across input views is associated with a 3D coordinate in a unified world frame, which inspires many variants and follow-ups \cite{slam3r,monst3r,must3r,fast3r,spann3r,cut3r,mast3r}.
WVD~\cite{WVD} leverages video diffusion to generate both videos and pointmaps jointly but is restricted to only static scenes.
Among them, MonST3R~\cite{monst3r} advocates a temporal extension, predicting \emph{4D pointmaps} for a pair of video frames from dynamic scenes. 
However, training such a method still demands many high-quality 4D data, which is hard to get~\cite{stereo4d}, and, in practice, depends on strong auxiliary modules~\cite{sam,sea_raft} plus lengthy post-optimization for global refinement.

In parallel, video diffusion models (VDMs)~\cite{ho2020denoising,rombach2022high,brooksvideo,chen2024gentron,girdhar2023emu}, trained at scale on vast unlabeled videos, have shown remarkable generative abilities, capturing not only the photometric properties of scenes but also coherent physical dynamics \cite{kondratyuk2023videopoet,menapace2024snap,sora,gen3c,genie2}. 
Recently, Marigold~\cite{marigold} has shown that DM could be repurposed for single image depth estimation by sharing a common latent space between the image and depth domains, while~\cite{depthcrafter,chronodepth} further extends the idea on VDMs but only for video depth estimation.
Building on this insight, such findings underscore a natural yet interesting question: \emph{can these general-purpose video diffusion backbones be leveraged to reconstruct full 4D geometry directly, obviating the need for massive annotated 4D datasets or cumbersome optimization pipelines?}

To this end, we affirm this question by proposing a novel approach, \emph{Sora3R}, to \emph{marry the learned dynamic “world knowledge” of video diffusion models~\cite{sora} with 4D pointmap representation.} 
Specifically, we introduce a two-stage framework that bridges the gap between pointmap regression and generative video models. 
First, we finetune a specialized \emph{pointmap VAE} from a pretrained video VAE to preserve latent-space compatibility, addressing the inherent distributional mismatch between video frames and 4D geometry. 
Then, we finetune a transformer-based diffusion backbone~\cite{opensora} within the combined latent space of video and pointmap.
The proposed \emph{Sora3R} is an efficient, feedforward framework that predicts coherent 4D pointmaps for all frames from a monocular video — \textbf{requiring no} external segmentation, optical flow, or iterative global alignment.
We summarize our contributions in three folds:
 \begin{itemize} 
 \item We adapt a pointmap VAE from a pretrained video VAE to encode latent 4D geometry while maintaining consistency for DiT learning.
 \item We present \textbf{Sora3R}, a novel video-diffusion pipeline that directly infers 4D pointmaps from a monocular video, enabling dynamic scene geometry reconstruction. 
 \item We demonstrate that Sora3R efficiently recovers both camera poses and dense scene structures without complex optimization, paving the way for many possible downstream 3D/4D tasks in dynamic real-world environments. \end{itemize}

%% file: figures/fig_train.tex
\begin{figure}[t]
    \centering
    \includegraphics[width=\linewidth]{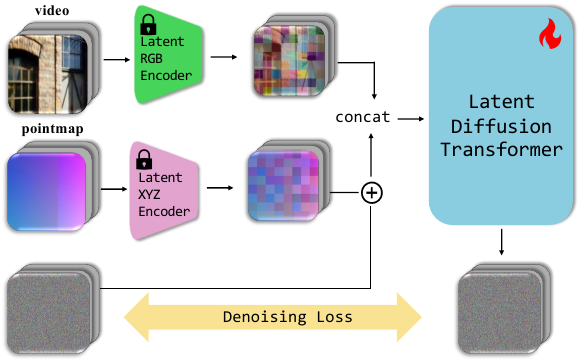}
    \caption{\textbf{Training Pipeline} During training, with pretrained video VAE encoder ($\mathcal{E}_{\text{RGB}}$) and pointmap VAE encoder ($\mathcal{E}_{\text{XYZ}}$), video latent and noisy pointmap latent are concatenated for training latent diffusion transformer with denoising loss.  }
    \label{fig:train}
\end{figure}

%% file: sec/2_related_works.tex
\section{Related works}

\mysection{Video Diffusion Models}
Diffusion models (DM) \cite{ho2020denoising,neal2001annealed,jarzynski1997equilibrium}, designed to recover data from noise incrementally, have been validated as scalable solutions for large generative systems, including multimodal image generation \cite{rombach2022high,dhariwal2021diffusion,ramesh2022hierarchical,saharia2022photorealistic,dai2023emu,xie2025sana,xie2024sana,chen2025janus,ma2024janusflow,fan2024fluid} and multimodal video generation \cite{chen2024gentron,brooksvideo,girdhar2023emu,menapace2024snap,kondratyuk2023videopoet,cong2023flatten,zhang2023show,blattmann2023align,esser2023structure,wang2023modelscope,gao2024lumina,wu2023tune,ho2022imagen,liu2024mardini,yin2024causvid,chen2025goku,kong2024hunyuanvideo}. In image generation tasks, the model produces an image aligned with the input text or class label. For video generation, temporal layers are added to the image-based DM, allowing the model to generate a video based on a text prompt, an image, or both.
Recent diffusion models primarily adopt the latent diffusion model architecture \cite{rombach2022high,blattmann2023align}, where a VAE-based compressor \cite{Kingma2013AutoEncodingVB,esser2021taming} embeds pixel-space values into a highly compressed latent code, and a diffusion model learns within this latent space. 
Empirical studies \cite{chen2024gentron, brooksvideo, girdhar2023emu,kondratyuk2023videopoet, wan21} indicate that, in this way, DM is able to learn from web-scale video data, and consequently demonstrate intriguing properties, such as capturing fundamental physical principles from videos. This has sparked interest in exploring the practical efficacy of video models in simulators \cite{brooksvideo,ha2018world,valevski2024diffusion,bruce2024genie}. 
While concurrent works~\cite{l4p_nvidia, scaling_4d_mae_deepmind} are studying the 
interplay between 3D/4D and video models through representation~\cite{gaussian_mae, gaussianvideo} and MAE~\cite{video_mae}, our work adopts the video diffusion model backbone, advancing beyond generative video methods to 4D reconstruction.

\mysection{3D and 4D Diffusion Models}
Generation tasks witness most of the application of 3D diffusion models~\cite{trellis, TripoSR_vast, sf3d_stability, cat3d, sv3d, clay, v3d, vfusion3d, see3d}, %
common approaches of which including SDS~\cite{dream_fusion, magic3d, reconfusion, richdreamer} optimization or
feed-forward inference~\cite{ one2345, one2345++, MVDream, llama_mesh, edgerunner_tang, lrm, gaussian_lrm}. 
Many of them~\cite{zero123, zero123plus, wonder3d, pose_free_lrm, im3d, magic123, reconx, zeronvs, viewcrafter, lvsm} are able to generate unseen parts given single-view or sparse-view observations.
Most recent dynamic 3D generation~\cite{cat4d, sv4d, stag4d, consistent4d} works have introduced the concept of 4D diffusion into the community.
Some of them inject camera pose condition~\cite{wonderland, ac3d, genie2, gen3c, cameractrl, camctrl3d, vd3d}, pointmaps~\cite{WVD}, object motion~\cite{freetraj, tc4d, 3dtrajmaster}, or point tracking~\cite{Go-with-the-Flow, motion_prompting} for controllable 4D consistency.
Existing literatures adopt a wide range of 4D representations, such as 4D NeRF or gaussian splatting~\cite{nerf, 3dgs, ar4d, l4gm, mav3d}, multiview videos~\cite{vividzoo, sync_cam_master, diffusion4d, 4diffusion}, or deformable geometry~\cite{dreammesh4d}.
Different from all of them, we adopt 4D pointmap from a reconstruction perspective as our representation.
Also, unlike concurrent DM works on reconstruction tasks to solely predict camera pose~\cite{ray_diffusion, PoseDiffusion, moe_llm_pose_diffusion}, depth~\cite{marigold, depthcrafter, chronodepth, rolling_depth}, or optical flow~\cite{scaling_marigold}, we predict 4D pointmap to reconstruct the full 4D geometry.

\mysection{3D and 4D Reoncstruction}
Classic vision-based methods for SLAM, SfM, and 3D reconstruction, based on multiview geometry~\cite{modern_ba,mvg_vgg} have been studied for decades.
Popular frameworks~\cite{past_present_future_slam, sfm_survey} like COLMAP~\cite{colmap} and ORB-SLAM~\cite{orb_slam, orb_slam3} have been the cornerstone for many downstream tasks and applications.
However, when extending to 4D reconstruction, they often filter out the dynamic components through motion segmentation~\cite{particlesfm, romo_sfm, dynaslam} to ensure geometry consistency, reconstructing static scenes only.
The bloom of learning-based methods brings many more new solutions for 3D and 4D reconstruction, especially dense mapping and reconstruction.
DROID-SLAM~\cite{deepv2d, droid_slam, megasam} predicts depth along with deep bundle adjustment.
Similarly, FlowMap~\cite{flowcam, flowmap} performs first-order optimization through vanilla gradient descent under optical flow guidance~\cite{raft}. 
Other methods~\cite{nerf--, sparf, tracknerf, nice_slam, glorie_slam} adopt view synthesis as the optimization objective while~\cite{pixelsplat, no_pose_no_problem} treat it as learning objective.
Feedforward methods~\cite{flare, pose_free_lrm, casualsam, geonet, nutworld, bullet_time, large_spatial_model} learning from different supervision predict reconstruction end-to-end.
Tracking-based methods~\cite{particlesfm, delta, vggsfm, splinegs, datap_sfm} like LEAP-VO~\cite{leapvo} learn reconstruction from static or dynamic point tracks.
ACE-Zero~\cite{ace_zero} re-formulates the reconstruction problem as scene coordinate regression.
Recently, DUSt3R~\cite{dust3r} inspires many follow-up ``3R'' works towards different directions, such as video depth estimation~\cite{align3r}, stateful reconstruction~\cite{spann3r, cut3r}, matching~\cite{mast3r}, SLAM/SfM~\cite{slam3r, mast3r_sfm, mast3r_slam}, visual localization~\cite{reloc3r}, multiview reconstruction~\cite{fast3r, mv_dust, must3r} and 4D reconstruction~\cite{monst3r, cut3r, fast3r, stereo4d}.
We focus on dynamic reconstruction with pointmap representation as MonST3R~\cite{monst3r}, but we adopt orthogonal video diffusion pipelines without lengthy global alignment and waive the dependencies on strong segmentation~\cite{sam} and optical flow~\cite{sea_raft} modules.

%% file: sec/4_method.tex
\section{Method}

\mysection{Overview} We describe our \emph{model design and training} in detail in Sec.~\ref{sec:xyz_vae} and Sec.~\ref{sec:4d_dit}, as shown in Fig.~\ref{fig:train}. 
We elaborate our \emph{model inference}, \textit{i.e.} the procedure for our model to generate 4D pointmaps, in Sec.~\ref{sec:inference} and Fig.~\ref{fig:inference}.
We illustrate our \emph{post-optimization}, \textit{i.e.} the process to infer intrinsic, extrinsic, and depth from 4D pointmaps, in Sec.~\ref{sec:post_optimization}.

\subsection{Temporal pointmap latent and VAE}\label{sec:xyz_vae}

Formally, given raw video frame input $\mathbf{V} \in \mathbb{R}^{N\times H\times W\times C}$, a typical pretrained temporal VAE with encoder $\mathcal{E}_{\text{RGB}}$ and decoder $\mathcal{D}_{\text{RGB}}$ learns to model the video latent distribution.

A pointmap~\cite{dust3r} is a group of pixel-wise 3D coordinates for a given frame, establishing the one-to-one pixel-point association in the world frame (usually the first frame).
Following the definition~\cite{dust3r} and notation~\cite{WVD} of pointmaps, while images have three RGB color channels, we denote the pointmap data as \textit{XYZ} since it has three spatial channels. 
We use interchangeable terms between temporal pointmaps and 4D (\textit{3D + temporal 1D}) pointmaps.

In contrast to existing approaches~\cite{marigold,depthcrafter,chronodepth,WVD} that freeze pretrained VAEs without additional tuning, we argue that fine-tuning is essential when transferring from temporal RGB images to temporal pointmaps. 
Real-world depth values can be extremely wide-ranging or effectively unbounded, causing the normalized pointmaps to become imbalanced, poorly scaled, and difficult for an unmodified video VAE to encode and decode.
To alleviate this input gap, we propose to learn the temporal pointmap latent that keeps close to video latent but has the ability to represent 4D geometry, as our implicit 4D representation for dynamic scene understanding.

In other words, our goal is to fine-tune RGB VAE $\{\mathcal{E}_{\text{RGB}},\mathcal{D}_{\text{RGB}}\}$ to get XYZ VAE model $\{\mathcal{E}_{\text{XYZ}},\mathcal{D}_{\text{XYZ}}\}$.
During the finetuning, we have known the groundtruth camera poses $\{\mathbf{T}_i\}_{i=1}^N$, where $\mathbf{T}_i \in \mathfrak{se}(3)$. 
We always perform preprocessing to guarantee that the first frame will be the coordinate frame, i.e., $\mathbf{T}_1=\mathbf{I}$. With depth maps $\mathbf{D} \in \mathbb{R}^{NHW}$ and known intrinsic $\mathbf{K}$ we can easily obtain the global pointmap, i.e., the temporal XYZ frame $\mathbf{P} \in \mathbb{R}^{NHWC}$, where:

\begin{equation}
\mathbf{P}_i(u, v) = \mathbf{T}_i \cdot \mathbf{K}^{-1} \begin{bmatrix}
u \cdot \mathbf{D}_i(u, v) \\ 
v \cdot \mathbf{D}_i(u, v) \\ 
\mathbf{D}_i(u, v)
\end{bmatrix}, \quad \forall i \in \{1, 2, \dots, N\}
    \label{eq:gt_xyz}
\end{equation}

We omit the homogeneous transform in Eq.~\ref{eq:gt_xyz} for brevity. 
Additionally, to normalize the wide-ranging scene coordinates, we apply a norm scale factor to the camera poses, depths, and pointmap. 
We set the norm factor as the average distance to the origin~\cite{dust3r}:

\begin{equation}
\mathbf{P}_i(u, v) = \frac{\mathbf{P}_i(u, v)}{\frac{1}{N \cdot H \cdot W} \sum_{i=1}^{N} \sum_{u=1}^{H} \sum_{v=1}^{W} \|\mathbf{P}_i(u, v)\|}
    \label{eq:scale_pointmap}
\end{equation}

However, we find that vanilla $L1$ reconstruction loss widely adopted in video VAEs lacks sensitivity to pointmap reconstruction precision, while $L2$ reconstruction loss will over-regularize the model on outlier points. Therefore, we reformulate Huber loss~\cite{stat_learning} as our XYZ VAE reconstruction loss:

\begin{equation}
\mathcal{L}_{\text{rec}}(\mathbf{\hat{P}}, \mathbf{P}) = 
\begin{cases} 
0.5 \cdot \|\mathbf{\hat{P}} - \mathbf{P}\|_2^2, & \text{if } \|\mathbf{\hat{P}} - \mathbf{P}\|_2 < \beta \\
\|\mathbf{\hat{P}} - \mathbf{P}\|_1 - 0.5 \cdot \beta, & \text{otherwise}
\end{cases}
\label{eq:vae_rec_loss}
\end{equation}

Note that $\mathcal{L}_{\text{rec}}$ is element-wise and only calculated on points with valid depths. Points in background regions like the sky with infinite depth will be masked. 
Finally, along with the standard Kullback-Leibler Divergence loss, we formulate the total training loss for our XYZ VAE as:

\begin{equation}
\mathcal{L}_{\{\mathcal{E}_{\text{XYZ}},\mathcal{D}_{\text{XYZ}}\}}=\mathcal{L}_{\text{rec}} + \lambda_{\text{KL}}\mathcal{L}_{\text{KL}}
\label{eq:vae_loss}
\end{equation}

\input{figures/fig_inference}

\subsection{4D Geometry DiT} \label{sec:4d_dit}

Once our XYZ VAE is trained, we now repurpose pretrained video diffusion models for denoising our proposed temporal pointmap latent, similar to canonical RGB video latent.
We adopt transformer-based architecture instead of UNet~\cite{rombach2022high} for its scalability and transferability by the spatial-temporal attention mechanism~\cite{dit_xie}.
Motivated by VDMs' emergent ability to understand world physics and dynamics~\cite{sora}, we start our fine-tuning with the model pretrained on massive web-scale videos to benefit from the strong spatiotemporal priors, mitigating the need for large-scale high-quality 4D annotated datasets which are hard to get~\cite{stereo4d, sync_cam_master}.

Inspired by~\cite{marigold}, as shown in Fig.~\ref{fig:train}, we first obtain RGB video latent $\textbf{H}_{RGB}$ and XYZ pointmap latent  $\textbf{H}_{XYZ}$ from $\mathcal{E}_{\text{RGB}}$ and $\mathcal{E}_{\text{XYZ}}$, respectively.
We train the latent diffusion transformer with rectified flow \cite{esser2024scaling, liu2022flow}.
For $\textbf{H}_{XYZ} = \mathcal{E}_{\text{XYZ}}(\textbf{P})$ and a normalized time step $t \in [0,1]$, the noised input $\textbf{H}_{XYZ}^t$ is sampled from a straight path between the target distribution and a standard normal distribution $\epsilon \sim \mathcal{N}(0,1)$:
\begin{equation}
    \textbf{H}_{XYZ}^t = t \textbf{H}_{XYZ} + (1-t) \epsilon.
\end{equation}
Then, we adopt the 4D DiT model $\mathcal{F}$ to predict the velocity $\boldsymbol{\nu}_\epsilon = \textbf{H}_{XYZ}^{t=1} - \textbf{H}_{XYZ}^{t=0}$, i.e., update $\mathcal{F}$ by minizing the learning objective:

\begin{equation}
    \mathbb{E}_{t,\mathbf{H}_{XYZ},\textbf{H}_{RGB},\epsilon} || \mathcal{F}(\textbf{H}_{XYZ}^t,\textbf{H}_{RGB},t)- \boldsymbol{\nu}_\epsilon||^2
\end{equation}
Here, $\textbf{H}_{RGB}$ is concatenated with $\textbf{H}_{XYZ}$,  serving as an additional condition for the denoising process.

We hypothesize that since XYZ VAE is fine-tuned from RGB VAE, though XYZ and RGB maps are very different ways to represent the 4D scene, in the hidden space, $\textbf{H}_{RGB}$ and $\textbf{H}_{XYZ}$ should share some internal features that can express the scene patterns.
As DiT was pretrained with RGB VAE before, as training on pointmap data proceeds, such conditioning can help DiT gradually adapt for denoising on 4D geometry latent domain through these bridging hidden representations.

\subsection{4D Pointmap Inference} \label{sec:inference}

During sampling, we only need $\mathcal{E}_{\text{RGB}}$ and $\mathcal{D}_{\text{XYZ}}$.
As shown in Fig.~\ref{fig:inference}, we sample random noise $\epsilon\sim \mathcal{N}(0,1)$ and concat it with video latent $\textbf{H}_{RGB}=\mathcal{E}_{\text{RGB}} (\textbf{V})$. 
The denoised temporal pointmap latent $\textbf{H}_{XYZ}$ from DiT can finally be decoded to 4D pointmaps $\mathbf{\hat{P}}=\mathcal{D}_{\text{XYZ}} (\textbf{H}_{XYZ})$.

The concatenated $\textbf{H}_{RGB}$ here not only serves as the denoising condition but also encodes rich spatiotemporal features from the video to help our DiT infer the 4D geometry.
Since our method processes all the video frames all at once through diffusion, we can capture the global spatiotemporal dependencies to infer temporal consistent, and spatial coherent 4D pointmaps.

\input{tables/datasets}
\input{tables/pose}

\subsection{Post-optimization for Downstream Tasks} \label{sec:post_optimization}
As we always fix the first frame as world coordinate frame during training, the predicted 4D pointmaps $\mathbf{\hat{P}}$ are supposed to share the same coordinate system too.
While $\mathbf{\hat{P}}$ itself can already serve as 4D representation of the scene, 
such representation makes it easy to support many geometry tasks with simple and straightforward post-optimization.

\mysection{Intrinsic Estimation} 
Since our goal is monocular video reconstruction, it's reasonable to assume that all the video frames coming from the same camera, \textit{i.e.}, share the same intrinsics. We set the principal point at the center of the frames. This implies: %

\begin{equation}
    c_x={W}/{2}, \quad c_y={H}/{2}
    \label{eq:pp_point}
\end{equation}

Since we fixed the first frame as the coordinate frame during training, we imply $\mathbf{\hat{T}}_1=\mathbf{I}$ by definition. 
According to Eq.~\ref{eq:gt_xyz}, we can solve the remaining focal $f$ for intrinsic $\mathbf{K}$ through optimization from $\mathbf{P}_1$ based on fast Weiszfeld algorithm~\cite{weiszfeld}:

\begin{equation}
\begin{split}
    \hat{f} &= \arg \min_{f} \sum_{u=1}^{W} \sum_{v=1}^{H} \\
    & \left\| (u-c_x, v-c_y)  - f \frac{(\mathbf{P}_1(u,v,0), \mathbf{P}_1(u,v,1)}{\mathbf{P}_1(u,v,2)} \right\|,
    \end{split}
    \label{eq:opt_K}
\end{equation}

\mysection{Camera Pose Estimation} 
After we obtain $\mathbf{\hat{T}}_1=\mathbf{I}$ and $\mathbf{\hat{K}}$ as above, we can easily infer the remaining camera poses from RANSAC PnP algorithm~\cite{mvg_vgg}:

\begin{equation}
\begin{split}
\mathbf{\hat{T}}_i &= \arg \min_{\mathbf{\hat{T_i}}} \sum_{u=1}^{W} \sum_{v=1}^{H} \left\| (u,v) - \pi \left( \mathbf{K} \mathbf{\hat{T}}_i \mathbf{P}_i(u,v) \right) \right\|_2 \\
& \text{subject to } \mathbf{\hat{T}}_i \in \mathfrak{se}(3), \quad \forall i \in \{1, 2, \dots, N\}
\end{split}
\label{eq:pnp_ransac}
\end{equation}

\mysection{Video Depth Estimation} 
Based on Eq.~\ref{eq:gt_xyz}, we can now easily get per-frame depth map estimation through simple pinhole projection: 
\begin{equation}
\mathbf{\hat{D}}_i = \mathbf{\hat{K}} \mathbf{\hat{T}}\mathbf{P}_i
\label{eq:opt_depth}
\end{equation}

%% file: figures/fig_inference.tex
\begin{figure}[t]
    \centering
    \includegraphics[width=\linewidth]{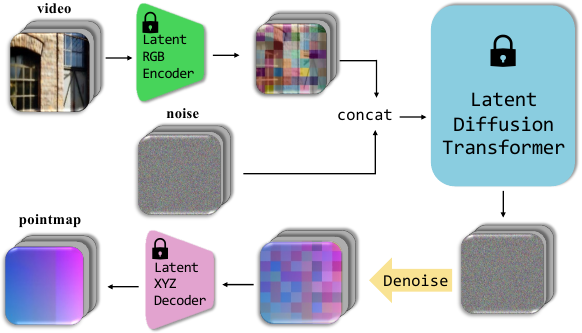}
    \caption{\textbf{Inference Pipeline} During testing, we sample random noise $\epsilon$ and concat it with video latent $\textbf{H}_{RGB}$. The denoised temporal pointmap latent $\textbf{H}_{XYZ}$ from DiT can finally be decoded to 4D pointmaps $\mathbf{\hat{P}}$ through $\mathcal{D}_{\text{XYZ}}$. }
    \label{fig:inference}
\end{figure}

%% file: tables/datasets.tex
\begin{table}[t!]
    \centering
    \resizebox{\columnwidth}{!}{%
        \begin{tabular}{l l l l l r r}
            \toprule
            \textbf{Name} & \textbf{Type} & \textbf{Camera Motion} & \textbf{Scene Motion} & \textbf{Sequences} & \textbf{Frames} \\
            \midrule
            DynamicReplica~\cite{dynamic_replica}  & Synthetic, Indoor          & Small     & Moderate & 483  & 144.9K  \\
            PointOdyssey~\cite{point_odyssey}    & Synthetic, Indoor \& Outdoor & Various   & Large    & 123  & 216.7K  \\
            Objaverse~\cite{objaverse}       & Synthetic, Objects         & Moderate  & Various  & 520  & 187.2K  \\
            TartanAir~\cite{tartanair}       & Synthetic, Indoor \& Outdoor & Large     & Static   & 369  & 306.6K  \\
            RealEstate10K~\cite{real10k}   & Realworld, Indoor \& Outdoor & Moderate  & Static   & 3753 & 671.8K  \\
            \bottomrule
        \end{tabular}%
    }
    \caption{\textbf{Training Data} We collect five diverse datasets spanning synthetic and real scenes with varying camera and scene motions. The total duration of the total $1.5B$ frames is around $14$ hours.}
    \label{tab:datasets}
\end{table}

%% file: tables/pose.tex
\begin{table*}[t]
\centering
\resizebox{1.8\columnwidth}{!}{%
\begin{tabular}{l ccc ccc ccc}
\toprule
 & \multicolumn{3}{c}{\textbf{Sintel}~\cite{sintel}} & \multicolumn{3}{c}{\textbf{TUM-dynamics}~\cite{tum_rgbd}} & \multicolumn{3}{c}{\textbf{ScanNet (static)}~\cite{scannet}} \\
\cmidrule(lr){2-4}\cmidrule(lr){5-7}\cmidrule(lr){8-10}
\textbf{Method}
& \textbf{ATE}\(\downarrow\) & \textbf{RPE trans}\(\downarrow\) & \textbf{RPE rot}\(\downarrow\)
& \textbf{ATE}\(\downarrow\) & \textbf{RPE trans}\(\downarrow\) & \textbf{RPE rot}\(\downarrow\)
& \textbf{ATE}\(\downarrow\) & \textbf{RPE trans}\(\downarrow\) & \textbf{RPE rot}\(\downarrow\) \\
\midrule
DUSt3R~\cite{dust3r}        & \cellcolor{yellow!40}0.320 & \cellcolor{orange!40}0.159 & \cellcolor{yellow!40}14.901  & \cellcolor{orange!40}0.026 & \cellcolor{orange!40}0.024 & \cellcolor{orange!40}1.102  & \cellcolor{orange!40}0.019 & \cellcolor{orange!40}0.023 & \cellcolor{orange!40}0.458 \\
MonST3R~\cite{monst3r}      & \cellcolor{red!40}0.155    & \cellcolor{red!40}0.064    & \cellcolor{red!40}1.739    & \cellcolor{red!40}0.009    & \cellcolor{red!40}0.010    & \cellcolor{red!40}0.725    & \cellcolor{red!40}0.011    & \cellcolor{red!40}0.011    & \cellcolor{red!40}0.315 \\
\midrule
Ours (Sora3r)& \cellcolor{orange!40}0.271  & \cellcolor{yellow!40}0.237 & \cellcolor{orange!40}9.437  & \cellcolor{yellow!40}0.029  & \cellcolor{yellow!40}0.028  & \cellcolor{yellow!40}2.386  & \cellcolor{yellow!40}0.040  & \cellcolor{yellow!40}0.042  & \cellcolor{yellow!40}2.348 \\
\bottomrule
\end{tabular}%
}
\caption{\textbf{Camera Pose Estimation Result}. Note different datasets have different depth scales, resulting in different numeric magnitudes for the metrics. We highlight the \colorbox{red!40}{1st}, 
\colorbox{orange!40}{2nd}, 
and \colorbox{yellow!40}{3rd} best results.}
\label{tab:pose}
\end{table*}

%% file: sec/5_experiment.tex
\section{Implementation Details}

\subsection{Datasets}
We present the training datasets in Tab.~\ref{tab:datasets}.
For real-world datasets, we use a subset of RealEstate10K~\cite{real10k} datasets. 
As RealEstate10K only has camera pose annotations, we run COLMAP~\cite{colmap} and DepthAnythingV2~\cite{depth_anything_v2} to get the aligned depth maps with the poses and then pointmaps.
Despite this, the resulting pointmaps remain somewhat noisy, so we filter the subset down to about $3,000$ sequences.
Because real-world datasets often yield incomplete or noisy depth and pose estimates, we bias our training heavily toward synthetic datasets, which can provide \emph{accurate and complete pointmap for every rendered frame}.
Specifically, we collect four synthetic datasets in total covering indoor and outdoor scenes with varying camera and scene motions: DynamicReplica~\cite{dynamic_replica}, Objaverse~\cite{objaverse}, PointOdyssey~\cite{point_odyssey}, and TartanAir~\cite{tartanair}.
For Objaverse, we randomly select $500$ objects spanning both dynamic and static categories, then render each with a random camera trajectory.

\subsection{Model Architecture} 
We build \emph{Sora3R} on top of OpenSora~\cite{opensora}. 
Specifically, we remove the discriminator and adversarial loss from the video VAE.
Our VAE compresses the video with a ratio of \(4\times8\times8\).
For the DiT, we remove the text encoder, text conditioning, and cross-attention. 
During initialization, we expand the DiT patchify layer by duplicating its weights to double the input channel dimension from 4 to 8, which allows us to tokenize $\textbf{H}_{RGB}$ and $\textbf{H}_{XYZ}$ jointly.
We use \texttt{bfloat16} precision for training and \texttt{float16} precision for inference.
We set the number of sampling steps as $100$ during the sampling at inference.

\subsection{Training Protocol}
We fix the spatiotemporal resolution as \(17\times384\times512\) for all video clips, with random temporal stride in $[1,3]$, following the common practice in video VAEs~\cite{magvit_v2}. 
Both our XYZ VAE and DiT models are trained in two stages. 
\textbf{First stage:} we warm up on RealEstate10K for $4$ epochs with a learning rate of \(1\times10^{-5}\). Although RealEstate10K data are large but somewhat noisy, it provides moderate camera motion in static scenes, making it suitable for initial pretraining.
\textbf{Second stage:} we fine-tune exclusively on synthetic datasets with a reduced learning rate of \(5\times10^{-6}\). 
The reason is that synthetic data offer perfect depth and camera poses, which ensure high-quality supervision for the final-stage finetuning. 
Each fine-tuning stage only takes about $48$ hours on $4$ Nvidia A100 GPUs.

\subsection{Evaluation Protocol}
We evaluate our Sora3R on three common but challenging unseen datasets in our training: Sintel~\cite{sintel}, TUM-dynamics~\cite{tum_rgbd}, and ScanNet~\cite{scannet}, which also spanning from static and dynamic scenes and covering from synthetic and real-world data with diverse camera motions.
For each dataset, we sample 50 sequences at uniform intervals, each with a \(17\times384\times512\) resolution and a temporal stride of 2.
Since there are no established metrics to evaluate 4D pointmaps, following MonST3R~\cite{monst3r}, we conduct camera pose estimation and video depth estimation to evaluate our 4D geometry.
For the depth evaluation, we apply the same scale and shift alignment used in MonST3R.
For the pose evaluation, we apply a \(\mathfrak{sim}(3)\) Umeyama alignment to match predictions with ground truth trajectories.
We also report the same metric as~\cite{monst3r}: Absolute Relative Error (Abs Rel), percentage of inlier points $\delta < 1.25$, Absolute Translation Error (ATE), Relative Translation Error (RPE trans), and Relative Rotation Error (RPE rot).

%% file: sec/5_result.tex
\input{figures/fig_comparisons}

\section{Result}

\subsection{Camera Pose Estimation}
As shown in Tab.~\ref{tab:pose}, overall, our \emph{Sora3R} demonstrates robust performance across all benchmarks but does not yet surpass the strongest baselines MonST3R, falling behind by a notable performance gap. 
For ScanNet (static), DUSt3R and MonST3R clearly lead the table, for instance, DUSt3R achieves $0.019$ ATE v.s. our $0.040$.
One reason is that DUSt3R trains on large-scale real-world indoor datasets like ScanNet++~\cite{scannet++}, whereas our real-world pretraining only involves noisy depth and poses from RealEstate10K.
Despite these gaps in static environments, \emph{Sora3R} remains competitive for dynamic cases. 
On Sintel, we outperform DUSt3R on ATE with $0.271$ v.s. $0.320$, on RPE rot. $9.437$ v.s. $14.901$.
We also achieve a comparable ATE on TUM-dynamics ($0.029$ v.s.\ $0.026$), although our overall performance there is not as strong as on Sintel, likely due to TUM-dynamics’ smaller human motion but extensive static real-world indoor layout. 
It's worth noting that both DUSt3R and MonST3R employ global alignment and pairwise graph optimizations, further refining their final pose accuracy.

\subsection{Video Depth Estimation}
Tab.~\ref{tab:video_depth} summarizes our video depth estimation results. 
ChronoDepth~\cite{chronodepth} and DepthCrafter~\cite{depthcrafter} have pipelines similar to ours but are dedicated to repurposing VDMs for video depth estimation.
Despite relying on PnP‐estimated poses from pointmaps (which introduces additional noise), we still outperform ChronoDepth and DepthCrafter in certain settings. 
For instance, on TUM‐dynamics, our method achieves an Abs Rel of $0.211$ v.s. $0.340$ and $0.269$, and a $\delta < 1.25$ of $0.742$ v.s. $0.446$ and $0.597$, respectively. 
We also see gains on ScanNet, where our $\delta < 1.25$ is $0.823$ v.s. $0.641$ and $0.688$, and our Abs Rel of $0.285$ is lower than DepthCrafter’s $0.344$.
We attribute these improvements to our fine‐tuning strategy, in which we adapt the video VAE specifically for understanding 4D geometry instead of relying on an unmodified video‐only VAE. 
Nevertheless, our approach still falls short of DUSt3R~\cite{dust3r} and MonST3R~\cite{monst3r}, especially in static scenes. 
One likely reason is their use of confidence‐map mechanisms that help avoid extreme depth values—such as those in background regions—resulting in smoother depth maps with lower variance (Fig.~\ref{fig:comparison}).

\input{tables/video_depth}

\subsection{Qualitative Results}
We visualize the recovered depth maps and camera trajectories in Fig.~\ref{fig:comparison}.
From the visualized depth maps, we find that our Sora3R tends to predict depth maps with wider value ranges, which actually align with the groundtruth distribution, while DUSt3R and MonST3R tend to predict smoother depth maps.
For the depth quality, though DUSt3R and MonST3R have shown better quality for static structures, our method also demonstrates the ability to reconstruct both scene structure and motions.
About the pose quality, though our pose recovered from 4D pointmaps by PnP is still noisy, it has understood the general camera motion.
The visualization verified that VDMs can indeed have a sense of the world coordinate, and be able to understand scene dynamics as well as camera motion.

%% file: figures/fig_comparisons.tex
\begin{figure*}[htb]
    \centering
    \includegraphics[width=\linewidth]{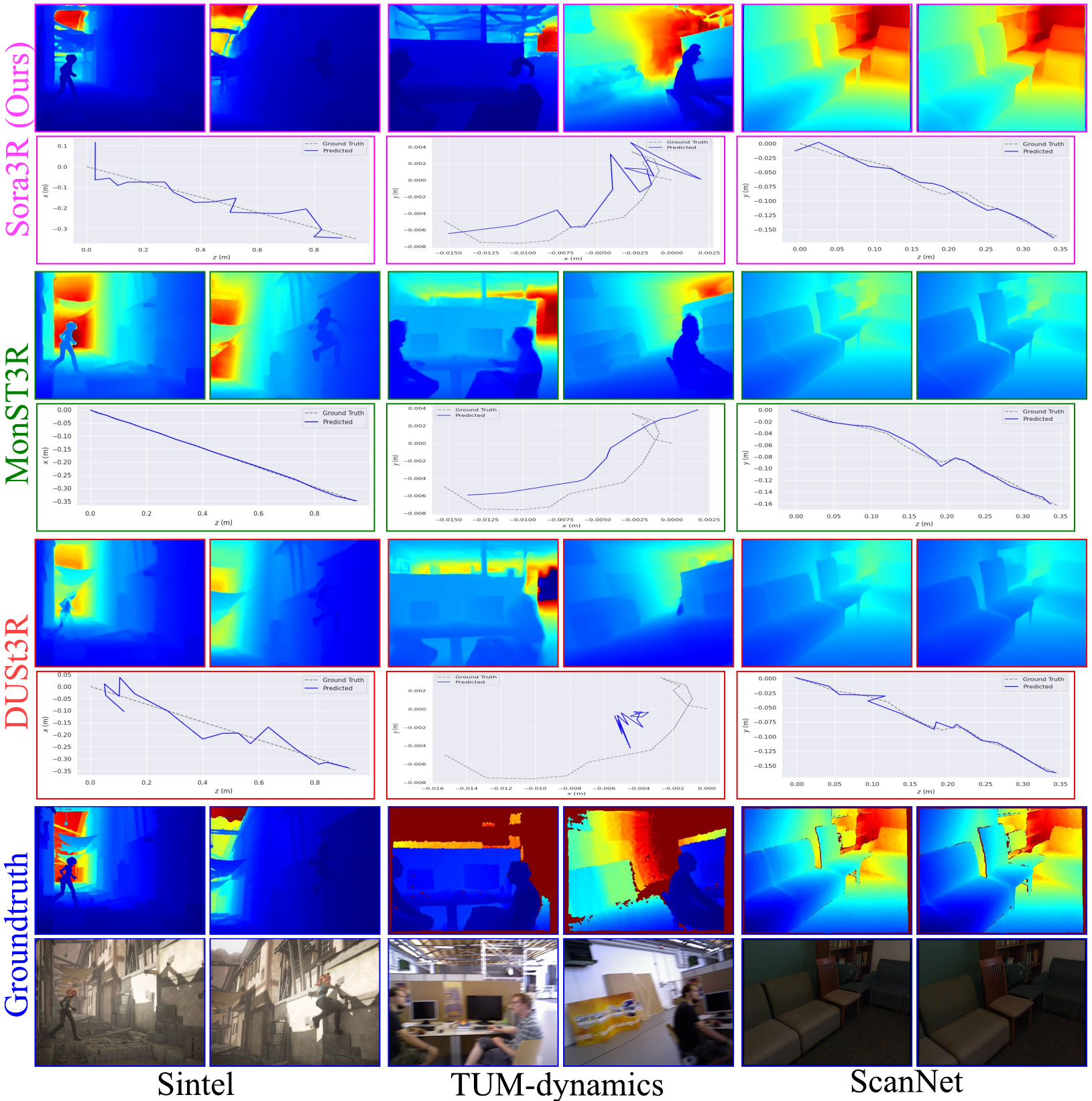}
    \caption{\textbf{Visualization Comparisons} From top to bottom is Sora3R (ours), MonST3R~\cite{monst3r}, DUSt3R~\cite{dust3r}, and groundtruth. For each method, the top row is the reconstructed depth map while the second row is the camera trajectory visualized together with the groundtruth trajectory. For groundtruth, the top row is depth while the second is the corresponding video frame. Since TUM-dynamics and ScanNet are obtained by depth cameras with missing or invalid pixels, they are marked with \colorbox{red!80}{dark red}.}
    \label{fig:comparison}
\end{figure*}

%% file: tables/video_depth.tex
\begin{table}[t!]
\centering
\resizebox{1\columnwidth}{!}{%
\begin{tabular}{lcccccc}
\toprule
\multirow{2}{*}{\textbf{Method}} 
& \multicolumn{2}{c}{\textbf{Sintel}~\cite{sintel}} 
& \multicolumn{2}{c}{\textbf{TUM-dynamics}~\cite{tum_rgbd}} 
& \multicolumn{2}{c}{\textbf{ScanNet (static)}~\cite{scannet}} \\
\cmidrule(lr){2-3}\cmidrule(lr){4-5}\cmidrule(lr){6-7}
& \textbf{Abs Rel} \(\downarrow\) & \(\delta < 1.25\) \(\uparrow\)
& \textbf{Abs Rel} \(\downarrow\) & \(\delta < 1.25\) \(\uparrow\)
& \textbf{Abs Rel} \(\downarrow\) & \(\delta < 1.25\) \(\uparrow\) \\
\midrule
ChronoDepth~\cite{chronodepth}   & \cellcolor{yellow!40}0.499 & 0.342  & 0.340  & 0.446  & \cellcolor{yellow!40}0.210 & 0.641 \\
DepthCrafter~\cite{depthcrafter}  & 1.282                  & \cellcolor{yellow!40}0.450 & 0.269  & 0.597  & 0.344                  & 0.688 \\
DUSt3R~\cite{dust3r}        & \cellcolor{orange!40}0.429 & \cellcolor{orange!40}0.539 & \cellcolor{orange!40}0.168 & \cellcolor{orange!40}0.808 & \cellcolor{red!40}0.026   & \cellcolor{red!40}0.990 \\
MonST3R~\cite{monst3r}       & \cellcolor{red!40}0.338   & \cellcolor{red!40}0.599   & \cellcolor{red!40}0.136   & \cellcolor{red!40}0.823   & \cellcolor{orange!40}0.031 & \cellcolor{orange!40}0.984 \\
Ours (Sora3r)& 0.544                  & 0.435  & \cellcolor{yellow!40}0.211 & \cellcolor{yellow!40}0.742 & 0.285                  & \cellcolor{yellow!40}0.823 \\
\bottomrule
\end{tabular}
}
\caption{\textbf{Video Depth Estimation Result} We highlight the \colorbox{red!40}{1st}, \colorbox{orange!40}{2nd}, and \colorbox{yellow!40}{3rd} best results.}
\label{tab:video_depth}
\end{table}

%% file: sec/6_ablations.tex
\input{figures/fig_runtime}

\input{tables/ablation}

\subsection{Ablation Study}

\subsubsection{Importance of 4D Pointmap Learning}
We conduct an extensive ablation study on the Sintel dataset in Tab.~\ref{tab:ablation} to analyze the effectiveness of our approach. 

In experiment (a), the pretrained video VAE$_{RGB}$, evaluated with ground-truth pointmaps, shows notably weaker performance than experiment (b), where VAE$_{RGB\rightarrow XYZ}$ is fine-tuned on pointmap data. 
A similar conclusion can be obtained by comparing experiment (c), fine-tuning DiT$_{RGB\rightarrow XYZ}$ on frozen video VAE$_{RGB}$, and experiment (e), jointly fine-tuning both the VAE$_{RGB\rightarrow XYZ}$ and the DiT$_{RGB\rightarrow XYZ}$ on pointmaps (RPE rot. $135.739$ v.s. $9.437$).
This considerable gap clearly illustrates that pretrained video VAEs, when remain frozen, struggle with encoding and decoding 4D pointmaps due to the inherent imbalance and scale variance. 
Our fine-tuning approach effectively bridges this gap, showing that 4D latent domain adaptation is crucial for learning accurate 4D geometry.

Furthermore, experiment (d), where both VAE$_{SCRATCH\rightarrow XYZ}$ and DiT$_{SRATCH\rightarrow XYZ}$ are trained from scratch on pointmaps, and (e) Sora3R, validates the central motivation behind our approach (Abs Rel $1.407$ v.s. $0.544$): the pretrained video latent representations encode valuable dynamic and geometric priors, and general-purpose video diffusion backbones can reconstruct 4D geometry with proper latent domain adaption and tuning. 

Interestingly, in experiment (b), VAE$_{RGB\rightarrow XYZ}$, which serves as a theoretical upper bound for our diffusion model, clearly outperforms the strongest baseline MonST3R.
We hope future research along this line can further narrow this performance gap.

\subsubsection{Inference Efficiency} 
We compare the inference efficiency in Fig.~\ref{fig:runtime}.
With spatiotemporal resolution as \(17\times384\times512\), we have a similar model inference time as MonST3R.
However, our model inference time could be easily further reduced by adopting video VAE with a higher compression ratio or decreasing sampling steps, while DuSt3R and MonST3R have to process multiple pairs depending on sequence length.
Additionally, since Sora3R predicts 4D pointmaps for all the video frames in the world frame all at once, we simply perform feedforward post-optimization to get pose and depth ($1.5s$ v.s. $\sim 30s$), while the baselines all need lengthy iterative global alignment and optimization, as their predicted pair-wise pointmaps are in different coordinate systems.

\subsection{Limitations}
The main concern of our method is that the performance still falls short of state-of-the-art 4D geometry reconstruction methods~\cite{monst3r}. 
A common failure case, for example, as shown in Fig.~\ref{fig:limitation}, when our model predicts imbalanced wide-ranging pointmaps, it will severely affect the robustness of camera pose estimation.
Another limitation is that the current model inference for a fixed number of frames in a temporal window.
However, we believe that by introducing proper regularization and scaling up training data with stronger video VAE and diffusion backbones, which can now generate minute-long videos~\cite{loong, sora}, reconstructing in \emph{4D latent space} with rich spatiotemporal priors can be an efficient and promising future direction to scale up 4D reconstruction for longer videos.

\input{figures/fig_limitation}

%% file: figures/fig_runtime.tex
\begin{figure}[t]
    \centering
    \includegraphics[width=\columnwidth]{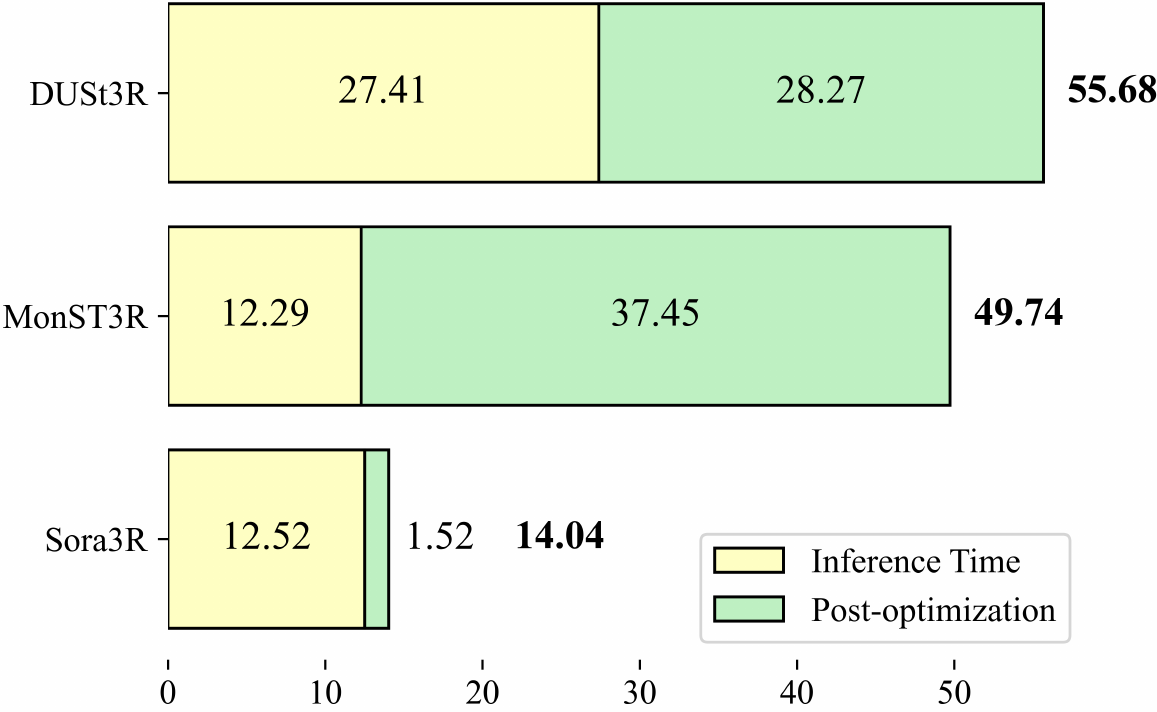}
    \caption{\textbf{Runtime Comparison} Average runtime (in seconds) to process video sequence with \(17\times384\times512\) spatiotemporal resolution, evaluated from $50$ runs on a single $A100$. }
    \label{fig:runtime}
\end{figure}

%% file: tables/ablation.tex
\begin{table*}[hbt!]
\centering
\resizebox{1.8\columnwidth}{!}{%
\begin{tabular}{llccccc}
\toprule
\multirow{2}{*}{\textbf{Method}} 
& \multirow{2}{*}{\textbf{Input}} 
& \multicolumn{3}{c}{\textbf{Pose}} 
& \multicolumn{2}{c}{\textbf{Video Depth}} \\
\cmidrule(lr){3-5}\cmidrule(lr){6-7}
& & \textbf{ATE} \(\downarrow\) & \textbf{RPE trans} \(\downarrow\) & \textbf{RPE rot} \(\downarrow\)
& \textbf{Abs Rel} \(\downarrow\) & \(\delta < 1.25\) \(\uparrow\) \\
\midrule
(a) VAE$_{RGB}$ 
    & GT Pointmap  
    & \cellcolor{orange!40}0.042 
    & \cellcolor{orange!40}0.036 
    & 34.223 
    & 0.584 
    & \cellcolor{orange!40}0.641 \\
(b) VAE$_{RGB \rightarrow XYZ}$ 
    & GT Pointmap  
    & \cellcolor{red!40}0.011 
    & \cellcolor{red!40}0.011 
    & \cellcolor{red!40}1.605 
    & \cellcolor{red!40}0.152 
    & \cellcolor{red!40}0.916 \\
\midrule
(c) VAE$_{RGB}$ + DiT$_{RGB \rightarrow XYZ}$ 
    & Video  
    & 0.267 
    & 0.099 
    & 135.739 
    & 1.407 
    & 0.116 \\
(d) VAE$_{SCRATCH\rightarrow XYZ}$ + DiT$_{SCRATCH\rightarrow XYZ}$       
    & Video     
    & 0.299 
    & 0.112 
    & 133.342 
    & 1.279 
    & 0.144 \\
\midrule
(e) VAE$_{RGB \rightarrow XYZ}$ + DiT$_{ RGB \rightarrow XYZ}$ (Sora3R) 
    & Video   
    & 0.271 
    & 0.237 
    & \cellcolor{yellow!40}9.437 
    & \cellcolor{yellow!40}0.544 
    & 0.435 \\
(f) MonST3R 
    & Video   
    & \cellcolor{yellow!40}0.155 
    & \cellcolor{yellow!40}0.064 
    & \cellcolor{orange!40}1.739 
    & \cellcolor{orange!40}0.338 
    & \cellcolor{yellow!40}0.599 \\
\bottomrule
\end{tabular}
}
\caption{\textbf{Ablation Study on Sintel Dataset.} The subscript $RGB$ denotes model pretrained on videos; $RGB\rightarrow XYZ$ denotes the model pretrained on videos and finetuned on pointmaps; $SCRATCH\rightarrow XYZ$ denotes the model initialized from scratch and trained on pointmaps. We highlight the \colorbox{red!40}{1st}, \colorbox{orange!40}{2nd}, and \colorbox{yellow!40}{3rd} best results.}
\label{tab:ablation}
\end{table*}

%% file: figures/fig_limitation.tex
\begin{figure}[htb!]
    \centering
    \includegraphics[width=\linewidth]{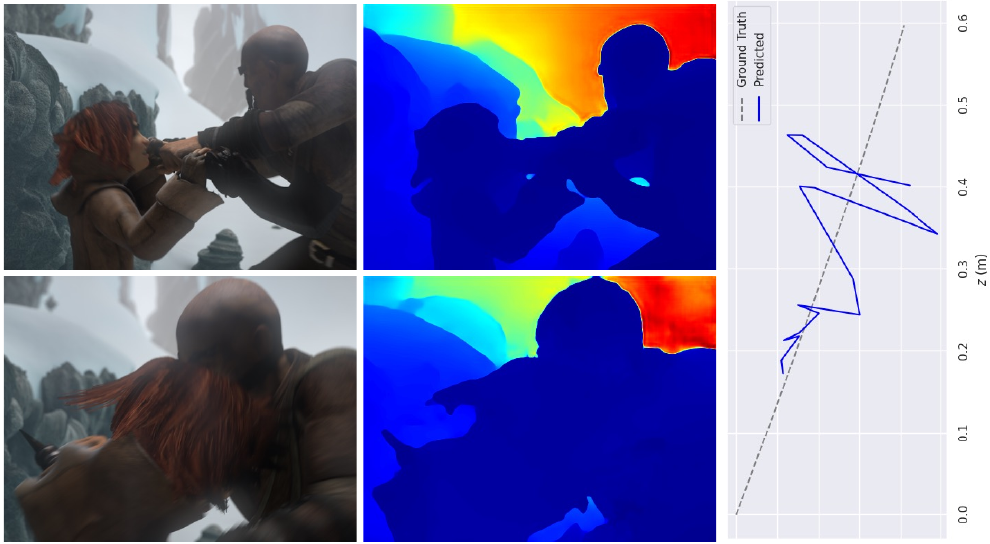}
    \caption{\textbf{Limitation: Example Failure Case.} \textit{Left: Video Frame; Middle: Recovered Depth; Right: Recovered Camera Trajectory.}  Sometimes, Sora3R predicts imbalanced pointclouds ranging from very near and very far, resulting in totally failed camera pose recovery.}
    \label{fig:limitation}
\end{figure}

%% file: sec/7_conclusion.tex
\section{Conclusion}

In this paper, we propose \emph{Sora3R}, a novel framework that repurposes video diffusion models (VDMs) for 4D geometry reconstruction from monocular video. 
By fine-tuning a specialized pointmap VAE and a transformer-based diffusion backbone, our method bridges the gap between pointmap regression and generative video modeling, effectively adapting the rich spatiotemporal latent priors for 4D reconstruction.
Although our current implementation exhibits a performance gap to state-of-the-art methods, through extensive experiments, we demonstrate that VDMs possess the “world knowledge” to capture physical dynamics and reconstruct dynamic 3D scenes in 4D pointmap latent space.
This demonstrates the potential of a generative model-based pipeline capable of handling both static and dynamic reconstruction without relying on external modules or sophisticated optimization.  
We hope that our work not only validates the potential of repurposing VDMs for 4D reconstruction but also inspires and catalyzes further research in dynamic scene reconstruction.

%% file: sec/8_acknowledgement.tex
\vspace{1em}

\mysection{Acknowledgement}
This publication was supported by funding from KAUST Center of Excellence on GenAI under award number 5940, as well as, the SDAIA-KAUST Center of Excellence in Data Science and Artificial Intelligence. 
For computer time, this research used Ibex managed by KAUST Supercomputing Core Laboratory.